\documentclass{article}

\usepackage{microtype}
\usepackage{graphicx}
\usepackage{subcaption}
\usepackage{booktabs} 

\usepackage{hyperref}

\usepackage{cuted}
\usepackage{lipsum}
\usepackage{xcolor}
\usepackage{listings}
\usepackage{caption} 

\usepackage[preprint]{icml2026}
\makeatletter
\let\ICMLorigPrintAffiliationsAndNotice\printAffiliationsAndNotice
\renewcommand{\printAffiliationsAndNotice}[1]{%
  \begingroup
  \renewcommand{\footnotetext}[2][]{}
  \let\@footnotetext\@gobble
  \ICMLorigPrintAffiliationsAndNotice{}%
  \endgroup
}
\makeatother

\usepackage{amsmath}
\usepackage{amssymb}
\usepackage{mathtools}
\usepackage{amsthm}
\usepackage{marvosym}

\usepackage[capitalize,noabbrev]{cleveref}

\theoremstyle{plain}

\theoremstyle{definition}

\theoremstyle{remark}

\usepackage[textsize=tiny]{todonotes}

\usepackage{cuted}
\usepackage{lipsum}
\usepackage{xcolor}
\usepackage{listings}
\usepackage{caption} 

\lstdefinelanguage{json}{
    basicstyle=\small\ttfamily, 
    numbers=left,
    numberstyle=\small\color{gray},
    stepnumber=1,
    numbersep=5pt,
    showstringspaces=false,
    breaklines=true,      
    breakatwhitespace=false, 
    frame=lines,
    backgroundcolor=\color{white},
    stringstyle=\color{red!60!black},
    keywordstyle=\color{blue},
    commentstyle=\color{gray},
    literate=
     *{0}{{{\color{magenta!60!black}0}}}{1}
      {1}{{{\color{magenta!60!black}1}}}{1}
      {2}{{{\color{magenta!60!black}2}}}{1}
      {3}{{{\color{magenta!60!black}3}}}{1}
      {4}{{{\color{magenta!60!black}4}}}{1}
      {5}{{{\color{magenta!60!black}5}}}{1}
      {6}{{{\color{magenta!60!black}6}}}{1}
      {7}{{{\color{magenta!60!black}7}}}{1}
      {8}{{{\color{magenta!60!black}8}}}{1}
      {9}{{{\color{magenta!60!black}9}}}{1}
      {:}{{{\color{red!60!black}{:}}}}{1}
      {,}{{{\color{red!60!black}{,}}}}{1}
      {\{}{{{\color{blue}{\{}}}}{1}
      {\}}{{{\color{blue}{\}}}}}{1}
      {[}{{{\color{blue}{[}}}}{1}
      {]}{{{\color{blue}{]}}}}{1},
}

\begin{document}

\twocolumn[
  \icmltitle{PhysForge: Generating Physics-Grounded 3D Assets for Interactive Virtual World}




  \begin{center}
    \textbf{Yunhan Yang}$^{*1,2}$, \textbf{Chunshi Wang}$^{*3,2}$, \textbf{Junliang Ye}$^{*4,2}$, \textbf{Yang Li}$^2$, \textbf{Zanxin Chen}$^5$, \\ \textbf{Zehuan Huang}$^6$, 
    \textbf{Yao Mu}$^5$, \textbf{Zhuo Chen}$^2$, \textbf{Chunchao Guo}$^{2}\textsuperscript{\Letter}$, \textbf{Xihui Liu}$^{1}\textsuperscript{\Letter}$
  \end{center}
  
  \begin{center}
    \footnotesize
    $^1$ HKU \quad
    $^2$ Tencent Hunyuan \quad
    $^3$ ZJU \quad
    $^4$ THU \quad
    $^5$ SJTU \quad
    $^6$ BUAA \quad
    \\
    \vspace{0.02in}
    $^*$ Equal Contribution \quad
    $\textsuperscript{\Letter}$ Corresponding Authors \quad
  \end{center}
  \vspace{0.08in}

  {\centering \tt\small \url{https://hku-mmlab.github.io/PhysForge/} \par}

  \vskip 0.3in
]



\printAffiliationsAndNotice{}  

\begin{strip}
    \centering
    \vspace{-3em}
    \includegraphics[width=\textwidth]{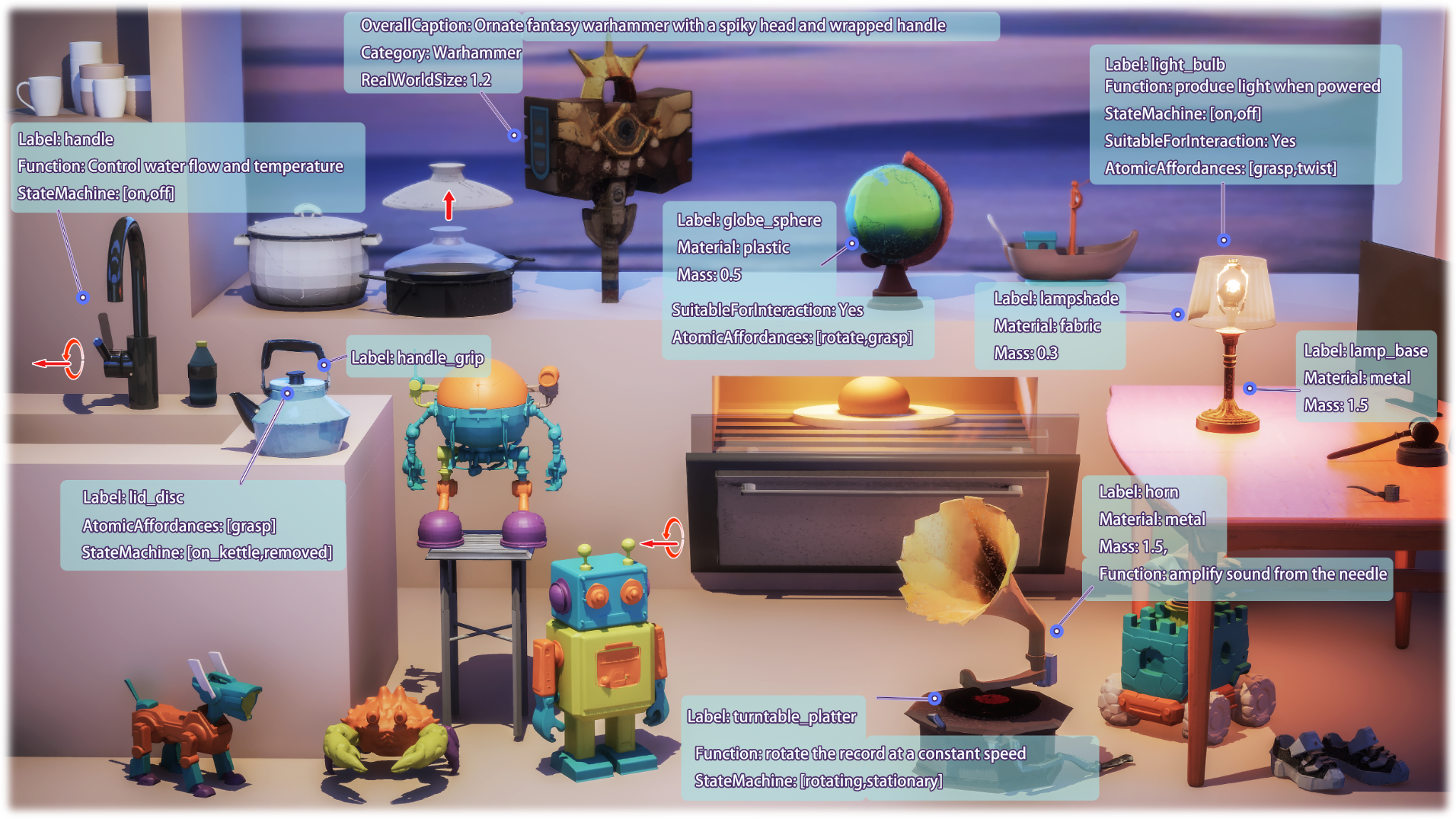}
    \captionof{figure}{PhysForge, takes only a single input image to generate Physics-Grounded 3D Assets. The figure showcases our high-quality generated results, where: (a) parts are distinguished by different colors to show their geometry; (b) parts are annotated with its detailed physical properties (text labels); and (c) the kinematic parameters for movable parts (such as joint axes) are precisely indicated by arrows. Our assets are ready for building Interactive Virtual World.}
    \label{fig:teaser}
\end{strip}

\begin{abstract}
Synthesizing physics-grounded 3D assets is a critical bottleneck for interactive virtual worlds and embodied AI. Existing methods predominantly focus on static geometry, overlooking the functional properties essential for interaction. We propose that interactive asset generation must be rooted in functional logic and hierarchical physics. To bridge this gap, we introduce PhysForge, a decoupled two-stage framework supported by PhysDB, a large-scale dataset of 150,000 assets with four-tier physical annotations. First, a VLM acts as a ``physical architect'' to plan a \textit{``Hierarchical Physical Blueprint"} defining material, functional, and kinematic constraints. Second, a physics-grounded diffusion model realizes this blueprint by synthesizing high-fidelity geometry alongside precise kinematic parameters via a novel \textit{KineVoxel Injection} (KVI) mechanism. Experiments demonstrate that PhysForge produces functionally plausible, simulation-ready assets, providing a robust data engine for interactive 3D content and embodied agents.
\end{abstract}    
\section{Introduction}
\label{sec:intro}
Recently, 3D generative models have achieved rapid progress, capable of synthesizing 3D assets with diverse appearances and high-fidelity geometric details~\cite{zhang20233dshape2vecset,xiang2024structured}. Concurrently, embodied AI and virtual game environments face a soaring demand for large-scale, high-quality 3D content. 3D generation technology holds the promise to serve as a \textit{data engine} to this content bottleneck. However, a significant gap remains: the vast majority of existing 3D generation methods focus solely on generating \textit{static} geometry and textures, overlooking the physics information that is crucial for interaction. These generated ``hollow shell'' assets cannot be grasped, pushed, or manipulated by agents, making them difficult to deploy directly in embodied AI simulators or game environments that require realistic physical interactions. To bridge this gap, we aim to propose a generation pipeline capable of producing \textit{physics-grounded} 3D assets directly. 

Our core insight is that for an object to be physically interactive, its generation must be driven by its \textit{functional logic} and \textit{hierarchical physics}. For example, a button on a television is the basic unit of function and operation; a cabinet's door and handle each carry distinct materials, functions, and kinematic definitions. Therefore, we shift the focus from traditional holistic shape generation to physics-centric synthesis, where the object's structure is a manifestation of its intended physical functions.

To achieve this, we propose \textbf{PhysForge}, an innovative two-stage framework that decouples physical planning from physical realization. Inspired by the ``planning-then-generation'' paradigms successful in 2D multimodal research~\cite{sun2024generative,chen2025blip3}, our design leverages the complementary strengths of specialized generative architectures: while VLMs possess the world knowledge necessary for complex physical planning, diffusion models excel at the precise synthesis of kinematic parameters, geometry, and textures. By decoupling these processes, PhysForge ensures that the generated assets are not only visually realistic but also physically consistent and simulation-ready.

The first stage is \textbf{VLM-based Planning}. Instead of starting from scratch, we finetune a powerful VLM, enabling it to acquire 3D spatial understanding and part-structure planning capabilities while retaining its inherent world knowledge. This VLM takes an image, an optional 2D mask, and generated 3D voxels~\cite{xiang2024structured} as input, and is tasked with generating what we call \textit{``Hierarchical Physical Blueprints''}. This blueprint includes the bounding box layout for all parts, as well as detailed physical properties for each part (including parent nodes, articulation types, etc.). We discover a critical synergistic effect: the introduction of physical properties, in turn, significantly aids the model's structural planning. By providing functional and physical constraints, it effectively resolves the ambiguity of part granularity, allowing the model to produce reasonable part decompositions even without 2D mask guidance.

The second stage is \textbf{Diffusion-based Generation}. After obtaining the \textit{blueprint}, we meticulously ``forge'' the high-fidelity geometry alongside the precise kinematic parameters promised in the planning stage.  We innovatively propose a KineVoxel Injection (KVI) mechanism. This method cleverly encodes precise articulation parameters (like origin, axis, and limit) into a special \textit{kinematic voxel}, allowing it to be jointly generated with the geometry-representing voxels during the diffusion denoising process-thereby achieving a synergistic synthesis of geometry and kinematic parameters.

To train our model effectively, we construct and introduce PhysDB, a large-scale dataset containing 150k assets. We define a novel four-tier annotation system that captures physics hierarchically. The holistic tier defines global properties like real-world scale and usage scene (e.g., kitchen, bedroom). The static properties tier covers part-level attributes such as semantic labels, physical materials (e.g., ``metal'', ``wood''), and mass. The functional tier defines part-level attributes such as intrinsic function (e.g., ``to contain'') and state machines (e.g., [open, closed]). Finally, the interactive tier specifies kinematic properties, including joint types (e.g., revolute, prismatic), and atomic affordances (e.g., pushable, graspable). 

PhysForge ultimately achieves the generation of functionally complete, physically interactive 3D assets from a single view image. Extensive experiments and qualitative demonstrations in physics simulator and game virtual world validate the effectiveness of the method, providing unprecedented high-fidelity, interactive assets for downstream applications such as robotic manipulation and game development.

Our core contributions are summarized as follows:
\begin{itemize}
    \item Formulation and Framework: We propose a novel formulation for physics-grounded 3D generation, and a decoupled \textit{VLM-based Planning + Diffusion-based Generation} two-stage framework (PhysForge).
    \item Large-scale Dataset: We contribute a large-scale, part-aware dataset with fine-grained, physical annotations (PhysDB), filling a critical data gap in the field.
    \item Extensive Validation and Application: We provide extensive experiments validating our framework's SOTA performance on both planning and generation, and demonstrate the direct applicability of our assets in robotic simulators and interactive virtual worlds.
\end{itemize}
\section{Related Work}
\label{sec:related}

\subsection{3D Content Generation}
The field of 3D content generation has rapidly expanded, largely following two distinct philosophies: leveraging powerful 2D priors or training directly on 3D data. A foundational strategy, Score Distillation Sampling (SDS) pioneered by DreamFusion~\cite{poole2022dreamfusion}, enables text-to-3D synthesis without 3D supervision by optimizing a 3D representation using gradients from a 2D model. 
This distillation paradigm was quickly adopted and improved upon by a vast body of work~\cite{wang2023score, wang2023prolificdreamer, lin2023magic3d, chen2023fantasia3d, metzer2022latentnerf, huang2023dreamtime, yi2023gaussiandreamer, wang2024animatabledreamer, wu2024consistent3d, alldieck2024score, tang2023stable, yan2024DreamView,ye2024dreamreward,liu2025dreamreward}.
Another line of work~\cite{liu2023syncdreamer, long2024wonder3d, shi2023zero123plus, liu2023one2345, liu2023one2345, yang2024dreamcomposer, xu2024instantmesh, qi2024tailor3d, zou2024triplane, huang2024epidiff, wen2025ouroboros3d} leverages 2D diffusion models to produce multi-view imagery, followed by reconstructing 3D geometry via multi-view consistency. To overcome the limitations of 2D priors, a distinct and growing body of research has focused on 3D-native generation. These methods train directly on large-scale 3D datasets, learning the underlying distribution of 3D shapes. The dominant approach in this area is latent diffusion, which requires a powerful 3D autoencoder to compress shapes into a manageable latent space. 
Significant progress has been made on 3D-native generation~\cite{zhao2023michelangelo,lai2025lattice,li2025triposg}, with models such as 3DShape2VecSet~\cite{zhang20233dshape2vecset} introducing an encoding scheme that uses cross-attention for set-structured 3D data, CLAY~\cite{zhang2024clay} scaling 3D diffusion to massive datasets, and TRELLIS~\cite{xiang2024structured} introducing structured latents for a high-quality, coarse-to-fine generation process. Despite this rapid evolution in synthesizing high-fidelity geometry and textures, a common limitation unites all these approaches: the resulting assets are holistic and non-interactive. 

\subsection{Part-aware 3D Shape Generation}
Recognizing the limitations of holistic generation, a recent line of work has begun to explore part-aware 3D generation~\cite{chen2024comboverse,liu2024part123,chen2024partgen,li2024pasta,yan2024phycage,tang2025efficient,lin2025partcrafter,tang2025efficient,yang2025omnipart,chen2025autopartgen,dong2025one,ding2025fullpart,he2025unipart}.
The central challenge in this sub-field is how to decompose a complex object into meaningful components while ensuring the final structure remains geometrically coherent. Early approaches have primarily adopted one of two strategies. The first is a ``reconstruction-from-views'' pipeline, which leverages 2D part masks to guide multi-view reconstruction~\cite{liu2024part123,chen2024partgen}. 
While this introduces part-level control, these methods often suffer from the same view-inconsistency issues as their holistic counterparts, resulting in low-fidelity geometry or parts that are merely surface-level segmentation rather than distinct objects. 
A significant advancement came from OmniPart~\cite{yang2025omnipart}, which introduced a two-stage framework built upon TRELLIS~\cite{xiang2024structured} to achieve semantic decoupling and structural cohesion, enabling controllable part generation. Other approaches, like PartPacker~\cite{tang2025efficient}, have focused on representation efficiency, compressing all parts into a compact dual volume representation for efficient generation from a single image. Critically, all these methods define \textit{parts} based on purely geometric or visual boundaries. Their goal is to create assets that are visually decomposable. This leaves a crucial gap: the function and physics of a part are never considered.

\subsection{Physics Grounded 3D Shape Generation}
Recently, a few pioneering works have begun to bridge the gap between static geometry and interactive physics. 
Some, like EmbodiedGen~\cite{wang2025embodiedgen}, have proposed comprehensive systems that integrate various generative modules, including layout generation, to create entire interactive scenes. 
PhysX-3D~\cite{cao2025physx3d} makes a significant contribution by introducing PhysXNet, a dataset annotating physical properties on top of PartNet~\cite{mo2019partnet}, and a generation model based on TRELLIS~\cite{xiang2024structured} using a Physical VAE. 
Separate from holistic physics, another body of research has focused specifically on articulation, a key component of interaction. This research has diverged into two main directions.
One specialized direction has concentrated on the reconstruction of articulated objects, often termed ``Digital Twins''~\cite{liu2023paris,liu2025building,weng2024neural,wu2025reartgs,song2024reacto,tu2025dreamo,cao2025physx}. 
A second direction attempts procedural generation of articulated assets~\cite{chen2024urdformer,gao2025meshart,le2024articulate,liu2024cage,liu2024singapo,mandi2024real2code,qiu2025articulate}. These approaches often rely on external, predefined content, such as part repositories, code templates, or VLM-predicted connectivity graphs, which constrains their ability to generalize to novel object categories and often leads to suboptimal accuracy.
\section{Physics-Grounded, Part-Aware 3D Assets Generation}
\label{sec:method_1}

\begin{figure*}
\centering
\includegraphics[width=\linewidth]{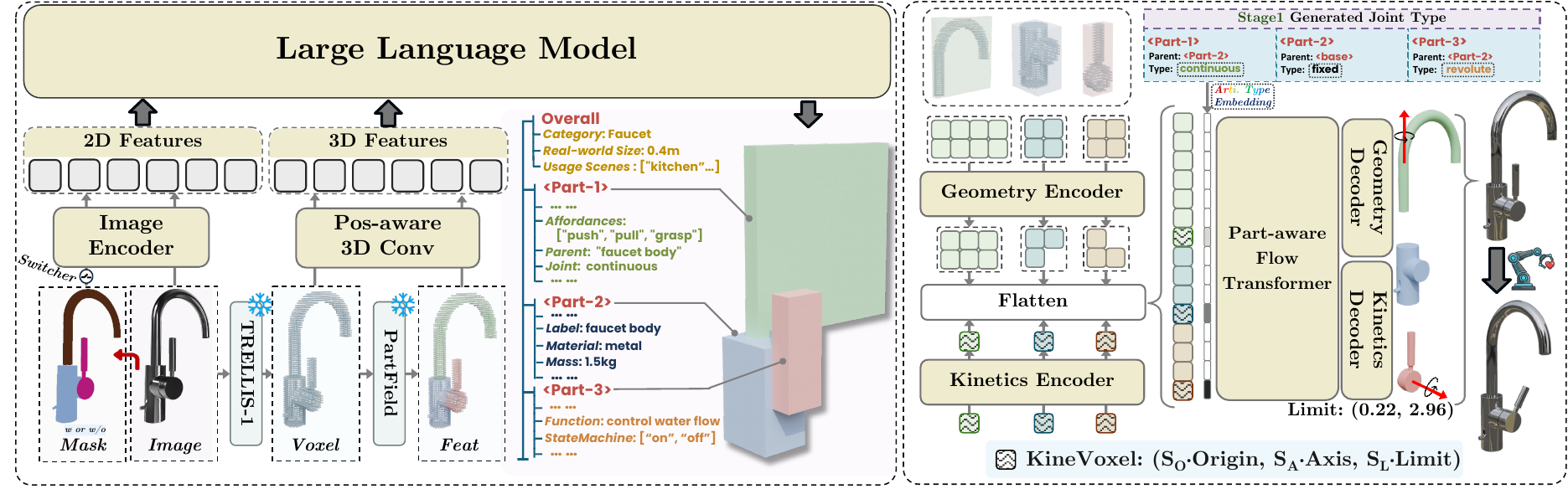}
\caption{\textbf{Method overview.} PhysForge consists of two stages: (Left) Stage 1: VLM-based Planning, where the VLM planner generates a ``Hierarchical Physical Blueprint" defining part structure and physical properties. (Right) Stage 2: Diffusion-based Generation, where a diffusion model, guided by the blueprint, uses the KineVoxel Injection (KVI) mechanism to synergistically generate the final geometry, texture, and precise kinematic parameters.}
\label{fig:pipeline}
\end{figure*}

Our goal is to generate physics-grounded 3D assets that can serve a wide range of domains, from embodied AI simulation environments to interactive video games. To achieve this, our approach is built upon two pillars: (1) a comprehensive and diverse training dataset, and (2) a powerful and robust generation pipeline.
We first introduce PhysDB, a novel large-scale dataset, in \Cref{sec:method_2}. It provides rich, fine-grained physical annotations necessary for this task.
Following this, we introduce a innovative two-stage generation framework PhysForge, as shown in \Cref{fig:pipeline}. Stage 1 (\Cref{sec:method_3}) is a ``VLM Planner'' that generates a hierarchical physical blueprint. Stage 2 (\Cref{sec:method_4}) is a ``Diffusion Realization'' stage, which uses a novel KineVoxel Injection mechanism to synthesize high-fidelity geometry, texture and precise articulation parameters.

\subsection{PhysDB: A Physics-Grounded Dataset}
\label{sec:method_2}
We propose a system of annotation that defines \textbf{holistic}, \textbf{static}, \textbf{functional}, and \textbf{interactive properties} to define the physical nature of each asset. At the object level, we define the asset's real-world scale, its object category, and its intended usage scene (e.g., kitchen, bedroom). Descending to the part level, we first define static and semantic properties, such as the part's semantic label, its physical material, and its mass. Next, we define functional properties inspired by OAKINK2~\cite{zhan2024oakink2}, which include the part's intrinsic function (e.g., ``to contain'', ``to control'') and its potential state machine (e.g., Button: [pressed, released]). Finally, our interactive tier specifies how an agent can interact with the object, detailing an atomic affordance library (e.g., pushable, rotatable) and, for movable parts, their complete kinematic definition: a parent part, a joint type (revolute, continuous, prismatic, or fixed), and the precise joint parameters (axis origin, direction, and limits).

We introduce PhysDB, a new dataset of 150k 3D objects sourced from Objaverse~\cite{deitke2022objaverse}, covering seven major categories: household, industrial, weapons, personal, vehicles, tech \& electronics, and cultural items. We select objects that are amenable to our physics annotation pipeline and already possess a meaningful part structure. 
Our annotation pipeline involves a human-in-the-loop process. We first render the whole objects and per-part images, which are fed to a multi modal LLM to generate initial annotations. This is followed by manual screening and correction to ensure the accuracy and consistency of the final PhysDB dataset.
Scaling precise 3D articulation annotation to 150k objects is extremely challenging. Due to the wide variety of object categories, PhysDB focuses on providing rich physical properties and identifying joint types, rather than attempting to annotate precise numerical axes which are often inaccurate at this scale. To bridge this kinematic gap, we supplement our training process with PartNet-Mobility~\cite{xiang2020sapien} and Infinite-Mobility~\cite{lian2025infinite}, which provide the ground-truth articulation parameters necessary to train our model in the diffusion stage.

\subsection{VLM as a Physical Blueprint Planner}
\label{sec:method_3}
The VLM's rich world knowledge provides a strong prior for object-part relationships, making it an ideal \textit{planner} for our first stage. While VLMs lack explicit 3D understanding, we finetune them to evoke this capability. We select Qwen2.5-VL~\cite{bai2025qwen2} as our base model due to its powerful knowledge base and vision capabilities.
To integrate 3D information, the model accepts a single image $I$, its corresponding 3D voxel representation $V$ (obtained from TRELLIS~\cite{xiang2024structured} first stage), and an optional 2D part mask $M$ for granularity control.
The input image $I$ and the 2D mask $M$ (which is converted to a color map) are processed directly by Qwen's powerful image encoder.
For the 3D voxel input $V$, we diverge from the common 3DShape2VecSet~\cite{zhang20233dshape2vecset} encoder. To better capture part-aware and local information, we first use a PartField encoder~\cite{liu2025partfield} to extract features for each voxel, then apply a position-aware 3D convolutional network to downsample these features into a 512-dimensional voxel embedding.

With these encoded inputs, We finetune the VLM to autoregressively generate the complete part structure and physical properties. We introduce 66 new special tokens to the VLM's codebook: \texttt{<boxs>} and \texttt{<boxe>} to delimit a bounding box, and 64 discrete tokens (\texttt{<box0>}, ..., \texttt{<box63>}) for the quantized coordinates. Each 3D axis-aligned bounding box is thus represented by only 6 tokens, enabling highly efficient structural planning. The model then outputs the hierarchical physical blueprint for each planned part.
A key discovery is that physics-guided planning resolves part ambiguity. Training the model to co-predict physical properties (like material and function) alongside bounding boxes provides stronger semantic constraints. This synergy significantly improves the model's understanding of part decomposition. As a result, even when no 2D mask is provided, the VLM can produce semantically coherent and reasonable bounding box plans.

\begin{table*}
\centering

\caption{Quantitative comparison of Physics Property generation on the PhysXNet. Our method outperforms the baseline in both geometric generation quality and the accuracy of predicted physical properties.}

\resizebox{0.95\linewidth}{!}{

\begin{tabular}{c|ccc|cccc}
\hline
 Method  & CD $\downarrow$ & F1-0.1 $\uparrow$ & F1-0.05 $\uparrow$ & Absolute scale (cm) $\downarrow$ & Material $\downarrow$ & Affordance $\downarrow$ & Description $\uparrow $ \\  
 \hline  
 TRELLIS & 10.10  & 86.53 &  72.47 & - &-&-&- \\
 PhysXGen & 9.81  & 87.91 & 73.60 & 25.83 & 1.59 & 3.69 & 0.38 \\
 PhysForge (Ours) & \textbf{9.21} & \textbf{89.24} & \textbf{75.43} &\textbf{11.04}& \textbf{0.81} & \textbf{1.22} & \textbf{0.87} \\
 
    \hline
\end{tabular}
 }
\label{tab:exp_physxnet}
\end{table*}
\begin{table*}
\centering

\caption{Quantitative comparison of Physics Property generation on the PhysDB. On our more diverse PhysDB dataset, our model demonstrates a more significant advantage over the baseline methods.}

\resizebox{0.95\linewidth}{!}{

\begin{tabular}{c|ccc|ccccc}
\hline
 Method  & CD $\downarrow$ & F1-0.1 $\uparrow$ & F1-0.05 $\uparrow$ & Absolute scale (m) $\downarrow$ & Material $\downarrow$ & Function $\uparrow$ & Interaction $\uparrow$ \\  
 \hline  
 TRELLIS & 24.32 & 68.19 & 53.28 &-&-&-&- \\
 PhysXGen & 25.30 & 65.79 & 50.57& 1.08 & 1.44 & 0.36 & 0.34 \\
 PhysForge (Ours) & \textbf{22.89}  & \textbf{70.51} & \textbf{55.38} & \textbf{0.37} & \textbf{0.43} & \textbf{0.83} & \textbf{0.96} \\
 
    \hline
\end{tabular}
}
\label{tab:exp_physics_ours}
\end{table*}
\begin{table}[!th]
\centering

\caption{Quantitative results for bounding box generation (\%) on PartObjaverse-Tiny. Results show that physics-guided planning significantly improves part planning accuracy and enables semantically reasonable results even without a 2D mask input.}

\resizebox{0.99\linewidth}{!}{

\begin{tabular}{c|ccc}
\hline
 Method  & Voxel recall $\uparrow$ & Voxel IoU $\uparrow$ & Bbox IoU $\uparrow$ \\  
 \hline  
 PartField & 69.65 & 46.04 & 37.33  \\
 OmniPart (SAM mask) & 68.33 & 43.34 & 34.33 \\
 PhysForge-bbox (w/o mask) & 67.89  & 35.53 & 32.30 \\ 
 PhysForge (w/o mask) & 73.63 & 47.66 &  36.32 \\ 
 OmniPart & 73.79 & 52.92 & 41.66  \\ 
 PhysForge (Ours) & \textbf{77.16} & \textbf{53.74} &  \textbf{42.95} \\ 
 
    \hline
\end{tabular}
}
\label{tab:exp_bbox_pot}
\end{table}

\subsection{Diffusion-based Generation with KineVoxel Injection}
\label{sec:method_4}
The VLM planner outputs a hierarchical structure, including per-part bounding boxes, parent-child relationships, and semantic joint types (e.g., fixed, revolute). While the VLM excels at this high-level structural and semantic planning, it is ill-suited for predicting the precise, continuous 3D values required for kinematics, such as an exact origin coordinate or axis vector. We therefore delegate this task to the diffusion head. This presents a challenge: how to synergistically generate these continuous parameters within a diffusion pipeline designed for geometry? 

We solve this by extending the OmniPart~\cite{yang2025omnipart} second stage framework with our novel KineVoxel Injection mechanism. 
Our approach begins by representing the articulation parameters for a single part $i$ as an 8-dimensional vector $P_i = (O_i, A_i, L_i)$, where $O_i \in \mathbb{R}^3$ is the joint origin, $A_i \in \mathbb{R}^3$ is the joint axis, and $L_i \in \mathbb{R}^2$ is the motion limits. We represent $P_i$ as a ``KineVoxel'', a special representation that can be processed alongside the standard geometric latents $Z_g$ in a unified denoising framework. Our approach maps data from different modalities (geometry and kinematics) into a unified latent space for joint diffusion. We utilize independent Kinematic Encoders ($E_{kine}$) and Decoders ($D_{kine}$) to process the KineVoxel, allowing it to share a latent space with the geometry latents within the middle transformer:
\[
z_{k,i} = E_{kine}(\text{concat}(S_O \cdot O_i, S_A \cdot A_i, S_L \cdot L_i)),
\]
where $S_O, S_A, S_L$ are scaling factors. Both $E_{kine}$ and $D_{kine}$ are implemented as lightweight 2-layer MLPs. The diffusion network contains down-sample blocks, a middle transformer, and up-sample blocks. We inject our KineVoxel $z_{k,i}$ after downsampling, concatenating it with the sequence of geometry voxel latents $Z_g = \{z_{g,i}\}$ before they are fed into the main denoising transformer. To allow the transformer to distinguish between the two latent types, we add a joint type embedding $E_{type}$ to the KineVoxel. This embedding $E_{type}$ is derived from the VLM's planned joint type (e.g., ``revolute'') and is added to $z_{k,i}$. The transformer can thus learn the complex correlations between part geometry and its corresponding joint parameters.

\begin{figure*}[!t]
\centering
\includegraphics[width=0.87\linewidth]{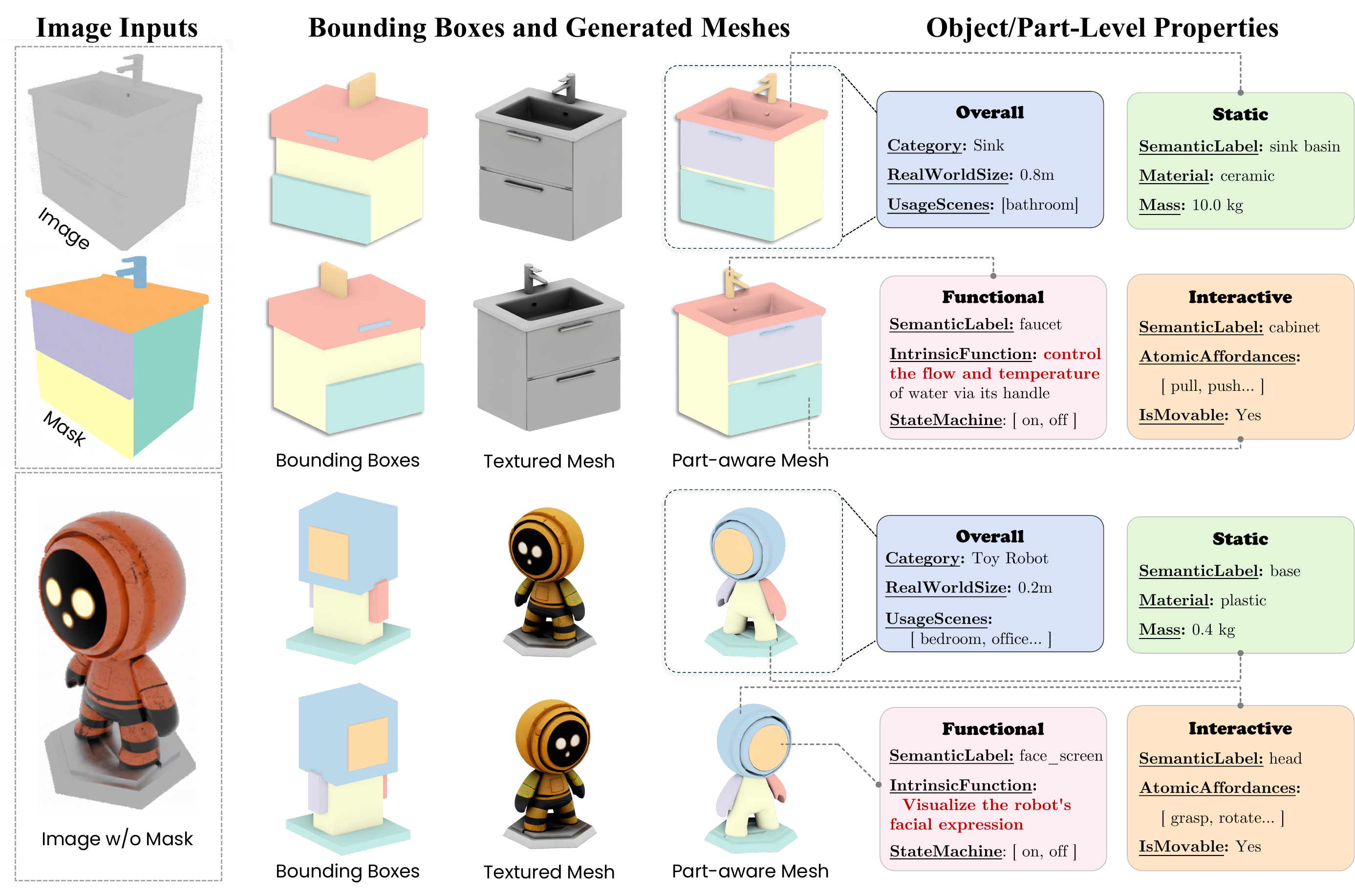}
\caption{Qualitative results of PhysForge. Given a single image and an optional 2D mask for control, our model generates high-quality, physics-grounded, and part-aware 3D assets.}
\label{fig:exp_ours_gen}
\end{figure*}

\begin{figure*}[!t]
\centering
\includegraphics[width=0.9\linewidth]{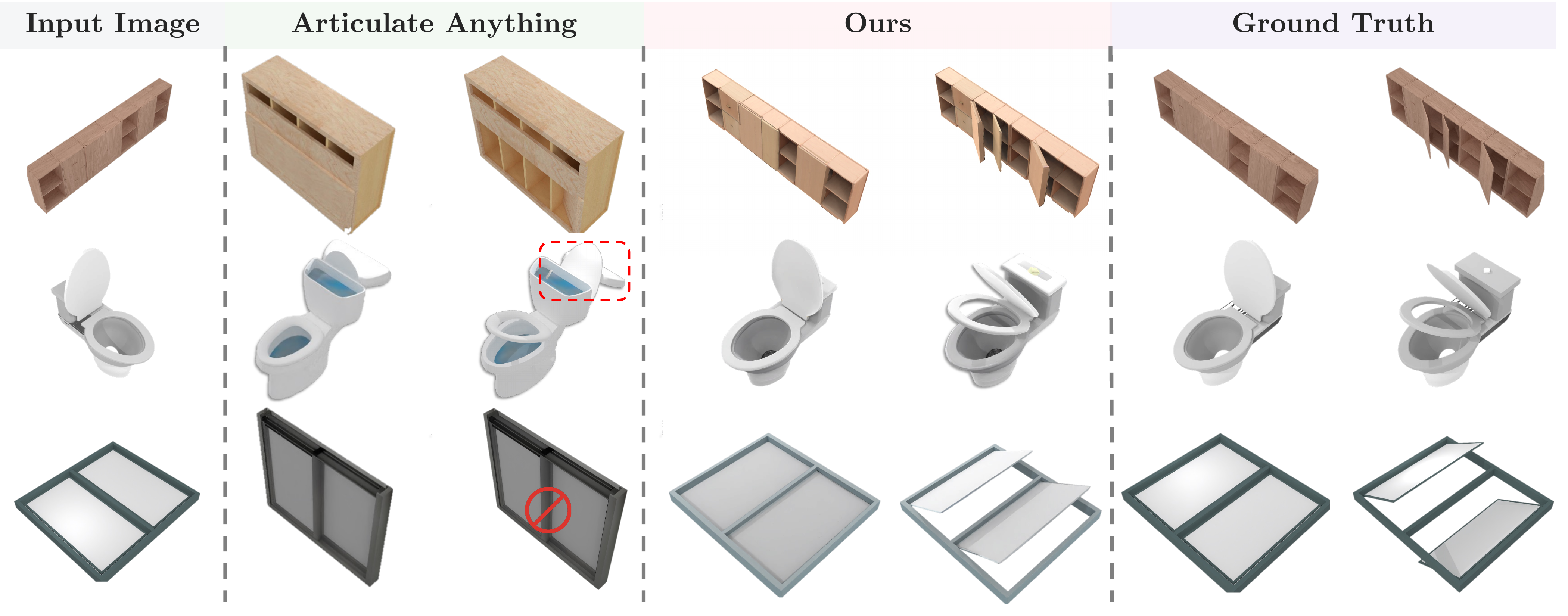}
\caption{Qualitative results of articulated object generation from a single image.}
\label{fig:exp_arti}
\end{figure*}

The entire model is trained by minimizing the Conditional Flow Matching (CFM) objective~\cite{lipman2022flow}. We define a composite loss that separates the contribution of geometry and kinematic voxels:
\[
\mathcal{L} = \mathbb{E}_{t, Z_0, c} \left[ \mathcal{L}_{geo} + \lambda_{kine} \cdot \mathcal{L}_{kine} \right]
\]
where $c$ is the condition from the VLM blueprint. The loss terms $\mathcal{L}_{geo}$ and $\mathcal{L}_{kine}$ are the standard $L_2$ losses between the predicted and target velocities for the geometry latents $Z_g$ and kinematic latents $Z_k$, respectively:
\[
\mathcal{L}_{geo} = \| v_{g,t} - \hat{v}_{g,t} \|^2; \mathcal{L}_{kine} = \| v_{k,t} - \hat{v}_{k,t} \|^2.
\]
We set the weighting factor $\lambda_{kine}=10$ throughout our training, placing a higher importance on accurately predicting the precise articulation parameters.
\section{Experiments}

\noindent\textbf{Evaluation Protocol.} 
To evaluate our model, we utilize the commonly used part-level dataset PartObjaverse-Tiny~\cite{yang2024sampart3d}, which contains 200 diverse objects, and the test set (1000 objects) from PhysXNet~\cite{cao2025physx}. We also establish two new test sets: (1) a set of 1,000 cases sampled uniformly by category from our proposed PhysDB, and (2) a set of 340 articulated objects sampled from PartNet- Mobility and Infinite-Mobility. We first evaluate our model's capability in the ``Part Structure Planning via VLM'' stage on the PartObjaverse-Tiny dataset, with results presented in \Cref{sec:exp_planning}. Following this, in \Cref{sec:exp_physics}, we evaluate the model's performance on generating accurate physical properties and kinematic parameters. Finally, We demonstrate the broad applications of our model in \Cref{sec:exp_physics}.

\subsection{Part Structure Planning}
\label{sec:exp_planning}
\noindent\textbf{Baselines and Metrics.}
We first evaluate and analyze our model's capability on the Part Structure Planning task. We select the first stage of OmniPart~\cite{yang2025omnipart} and PartField~\cite{liu2025partfield} as our primary baselines. The first stage of OmniPart stage trains an auto-regressive transformer on part-level data for bounding box generation, which, by default, requires a 2D mask input to control the granularity of the generated parts. PartField is a point cloud segmentation method that can also take voxels as input to produce voxel-level segmentation results and corresponding bounding boxes. As PartField requires the number of parts to define the segmentation scale, we provide the ground-truth number of parts as input. Following OmniPart, we use BBox IoU, Voxel Recall, and Voxel IoU as our evaluation metrics, assessing both bounding box-level accuracy and voxel-level planning precision.

\begin{figure*}[!t]
\centering
\includegraphics[width=0.9\linewidth]{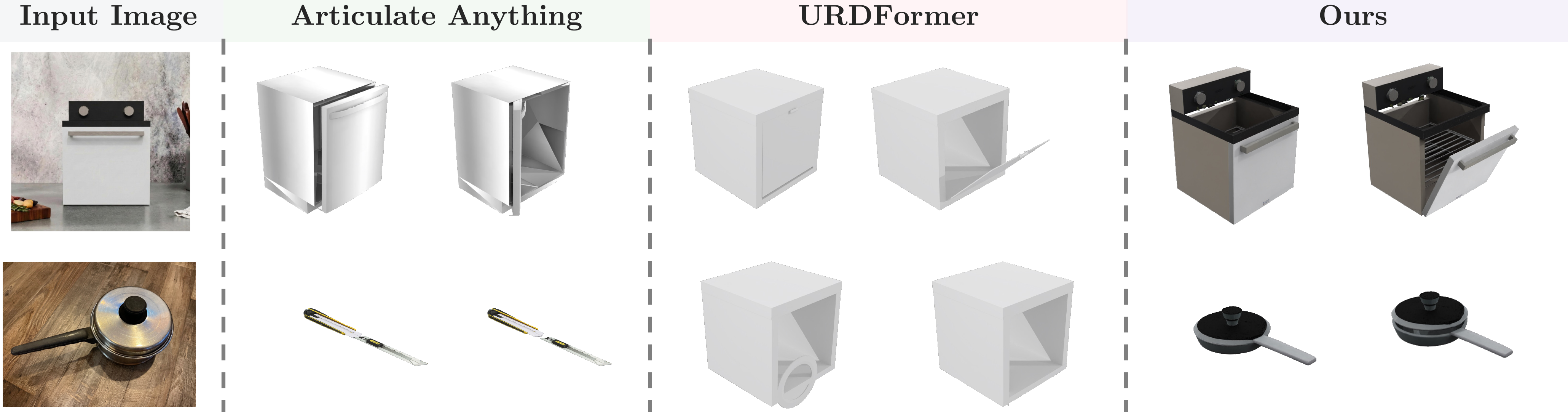}
\caption{Qualitative results of articulated object generation from a in-the-wild image.}
\label{fig:arti_wild}
\end{figure*}

\noindent\textbf{Results and Ablation Analysis.}
In \Cref{tab:exp_bbox_pot}, we show the comparison of all methods on the part structure planning task. To analyze our model's planning ability with and without a 2D mask input, we introduce two additional experimental settings. The second row, ``OmniPart (SAM mask)'', replaces OmniPart's ground-truth 2D mask with a 2D mask obtained from SAM~\cite{kirillov2023segment}, filtering out small masks with an area ratio less than $1600/1024^2$. The third row, ``PhysForge-bbox'', represents our model architecture trained only on the 500k part-level bounding box dataset (without physics). An entry marked ``w/o mask'' indicates that no mask was provided to the model input.

Comparing the overall results, our full model achieves state-of-the-art results, demonstrating the strongest part structure planning capability. The results of ``PhysForge w/o mask'' (row 4) are significantly better than the ``PhysForge-bbox'' model (row 3), which demonstrates that the introduction of physical properties significantly enhances our model's semantic understanding and planning capabilities for part structures. Even without a mask input, it can still produce semantically reasonable results. Furthermore, our model operating without a mask still outperforms the OmniPart's first stage that uses SAM-generated masks, highlighting the robustness of our physics-guided planning.

\begin{table*}[!t]
\centering

\caption{Quantitative comparison of articulated objects generation. Our method achieves higher fidelity to the input image and more accurate joint axis and pivot prediction.}

\resizebox{0.95\linewidth}{!}{

\begin{tabular}{c|cc|cc|cc}
\hline
 Method  & CD $\downarrow$ & Clip-Sim $\uparrow$ & Joint-Axis-Err-5 $\downarrow$ & Joint-Pivot-Err-5 $\downarrow$ & Joint-Axis-Err-all $\downarrow$ & Joint-Pivot-Err-all $\downarrow$ \\  
 \hline  
 Articulate Anything & 23.31 & 0.87 & 0.608 & 0.257 & 0.694 & 0.197 \\ 
 Singapo & 21.10 & 0.85 & 0.241 & 0.153 & - & - \\
 URDFormer & 25.42 & 0.84 & 0.781 & 0.652 & - & -  \\
 PhysForge (w/o joint type emb) & 10.73 & 0.90 & 0.157 & 0.132 & 0.292 & 0.141  \\ 
 PhysForge (w/o kinetics enc) & 11.31 & 0.89 & 0.158 & 0.117 & 0.204 & 0.120 \\ 
 \textbf{PhysForge} (Ours) & \textbf{10.21} & \textbf{0.93} & \textbf{0.101} & \textbf{0.071} & \textbf{0.164} & \textbf{0.096}  \\ 
 
    \hline
\end{tabular}
}
\label{tab:exp_articulation}
\end{table*}

\begin{figure*}[!t]
\centering
\includegraphics[width=0.9\linewidth]{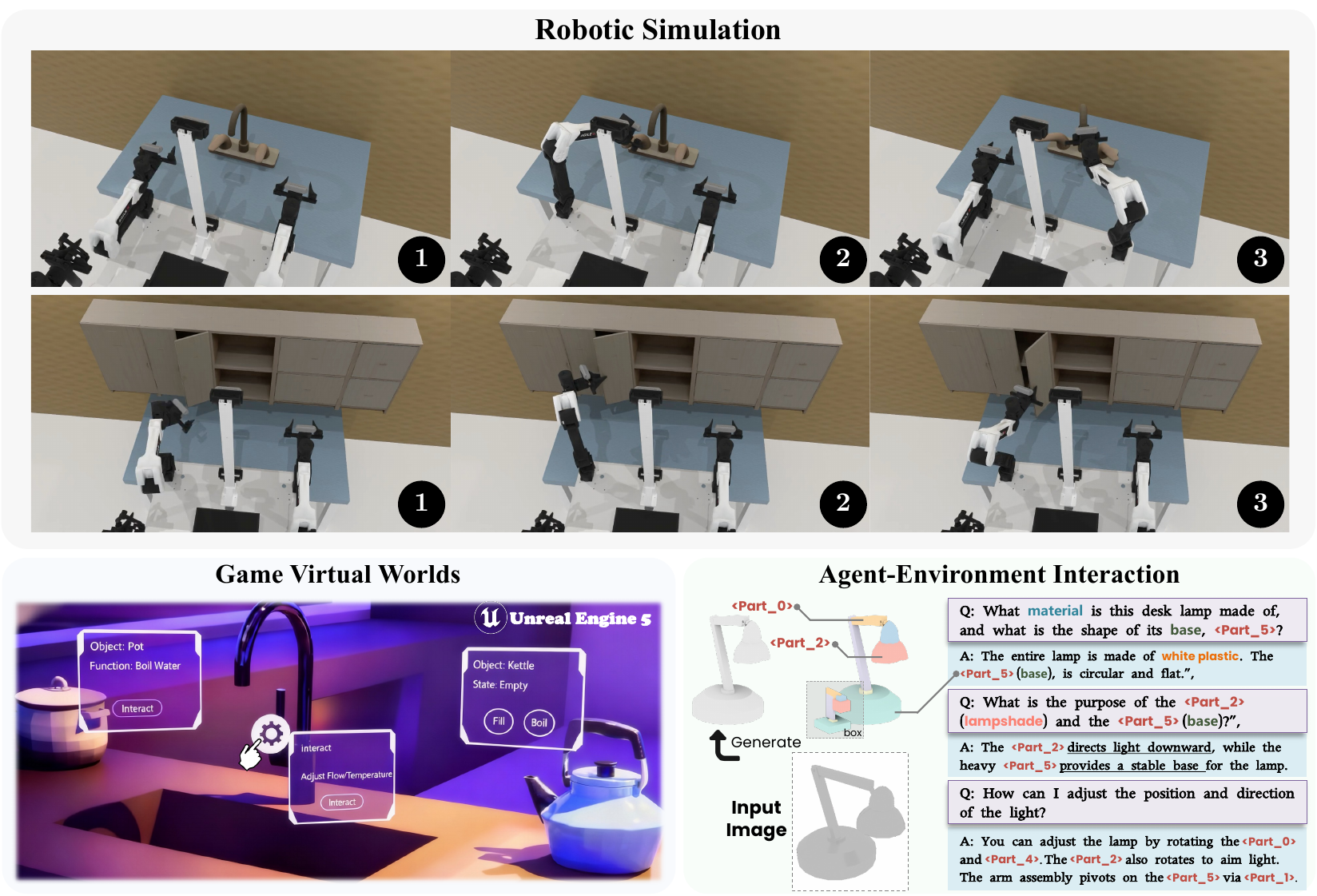}
\caption{\textbf{Downstream Applications of PhysForge.} Our generated assets are simulation-ready: (a) A robotic arm manipulates an asset's functional parts in a RoboTwin~\cite{Mu_2025_CVPR,chen2025robotwin} simulator. (b) The assets are imported into a virtual world (e.g., Unity/UE), enabling rich, physics-based interactions. (c) An agent interacts with our model via natural language, querying its physical blueprint to plan a task.}
\label{fig:application}
\end{figure*}

\subsection{Physics-Grounded Generation}
\label{sec:exp_physics}
\noindent\textbf{Baselines and Metrics.}
We evaluate our model's ability to generate physics properties by selecting PhysXGen and TRELLIS as our primary baseline. 
Specifically, we normalize the ground-truth and predicted shapes into a canonical space of [-0.5,0.5], then compute the Chamfer Distance (CD) and F1-Score. The F1-Score is assessed at two distance thresholds, $CD<0.1$ and $CD<0.05$. To evaluate the accuracy of physics properties at the part level, we compare the MAE of Absolute Scale, Material, Affordance, and the CLIP-Similarity of text-based Function and Interaction.

To evaluate our model's performance in generating Kinematic Parameters, we select Articulate Anything~\cite{le2024articulate}, Singapo and URDFormer as baselines, as they support articulated object generation from a single image. 
For this task, we use CD (\%) and F-Score (\%) to measure mesh generation quality, and CLIP-Similarity to assess the match with the input image. Following Articulate Anything, we utilize Joint Axis Error and Joint Pivot Error to measure the accuracy of the generated kinematic parameters. Specifically, we report Joint-Axis-Err-5 and Joint-Pivot-Err-5 on the subset of categories supported by all methods, and additionally report Joint-Axis-Err-all and Joint-Pivot-Err-all for methods that generalize to all categories.

\noindent\textbf{Physics Properties.}
In \Cref{tab:exp_physxnet}, we present a comparison of our method against the baselines PhysXGen and TRELLIS. Our method surpasses the other methods in terms of geometry generation quality. Unlike PhysXGen, which is trained on specific categories and is limited to outputting opaque CLIP features, our method benefits from the VLM's powerful world-knowledge prior, enabling it to directly and accurately output corresponding physics properties as both text and numerical values. Therefore, in the realm of physics property generation, our method significantly outperforms the baseline. Furthermore, in \Cref{fig:exp_ours_gen}, we demonstrate our model's effectiveness in generating physics-grounded assets. From a single image, along with optional 2D mask control, our pipeline can accurately plan all part bounding boxes and physical attributes, and subsequently utilize the diffusion model to generate part-aware geometry and textures.

\noindent\textbf{Kinematic Parameters.}
In \Cref{tab:exp_articulation}, we present the quantitative comparison between our method and the baseline models, along with qualitative results on the validation set (\Cref{fig:exp_arti}) and on in-the-wild images (\Cref{fig:arti_wild}). The articulated objects generated by our method are significantly superior to the baseline in terms of both consistency with the input image and the accuracy of the joint parameters.

\noindent\textbf{Ablation Analysis.}
We report the results of two key ablation studies in \cref{tab:exp_articulation}, which analyze the impact of removing the joint type embedding and the dedicated kinematic sub-network. The joint type embedding serves as the critical interface between our two stages: while Stage 1 predicts the qualitative articulation type (e.g., revolute, prismatic), this embedding provides a strong functional prior that constrains and guides the precise parameter estimation in Stage 2.
The results clearly demonstrate that without the guidance from Stage 1's planning, Stage 2 struggles to resolve kinematic ambiguities, leading to a degradation in joint accuracy, confirming that the joint type embedding is indispensable for effectively transferring physical common sense to the generation stage. Furthermore, removing the independent kinematic encoder and decoder further compromises the model's ability to synthesize precise mechanical constraints.

\subsection{Application}
To demonstrate the downstream utility of our generated assets, we showcase three primary applications in \Cref{fig:application}:
(a) Robotic Simulation: We demonstrate that our generated assets can be successfully imported into the RoboTwin~\cite{Mu_2025_CVPR,chen2025robotwin} simulation environment. The detailed part-level geometry and precise kinematic parameters allow robotic manipulators to realistically interact with the objects.
(b) Virtual Worlds: In game engines and virtual worlds, our assets enable complex interactions. Because every part is endowed with physics-grounded attributes (materials, mass, articulation), developers can design sophisticated interaction logic without manual rigging.
(c) Agent-Environment Interaction: Our VLM-based framework opens a new modality for interaction. An embodied agent (or VLA) can directly query our model in natural language and receive a text-based physical blueprint with bounding boxes, providing an explicit plan for manipulation.

\section{Conclusion}
We introduce PhysForge, a novel framework that generates interactive and physics-grounded 3D assets. Our decoupled ``VLM Planning + Diffusion Realization'' architecture finetunes a VLM to generate ``Hierarchical Physical Blueprints" that define an asset's complete physical profile. To realize these blueprints, our KineVoxel Injection algorithm enables a diffusion model to synergistically generate geometry and precise kinematic parameters. This framework is supported by PhysDB, our large-scale, 150k-asset dataset with rich annotations. PhysForge provides a foundational data engine for embodied AI and interactive virtual worlds.

\bibliographystyle{icml2026}
\bibliography{example_paper}

@String(CVPR= {IEEE Conf. Comput. Vis. Pattern Recog.})

@String(ICCV= {Int. Conf. Comput. Vis.})

@String(ECCV= {Eur. Conf. Comput. Vis.})

@String(TOG= {ACM Trans. Graph.})

@String(ICLR = {Int. Conf. Learn. Represent.})

@String(CVPR  = {CVPR})

@String(ICCV  = {ICCV})

@String(ECCV  = {ECCV})

@String(TOG   = {ACM TOG})

@String(ICLR  = {ICLR})

@article{zhang2024clay,
  title={CLAY: A Controllable Large-scale Generative Model for Creating High-quality 3D Assets},
  author={Zhang, Longwen and Wang, Ziyu and Zhang, Qixuan and Qiu, Qiwei and Pang, Anqi and Jiang, Haoran and Yang, Wei and Xu, Lan and Yu, Jingyi},
  journal={ACM Transactions on Graphics (TOG)},
  volume={43},
  number={4},
  pages={1--20},
  year={2024},
}

@inproceedings{poole2022dreamfusion,
  title={DreamFusion: Text-to-3D using 2D Diffusion},
  author={Poole, Ben and Jain, Ajay and Barron, Jonathan T and Mildenhall, Ben},
  booktitle={ICLR},
  year={2023}
}

@article{zhang20233dshape2vecset,
  title={3dshape2vecset: A 3d shape representation for neural fields and generative diffusion models},
  author={Zhang, Biao and Tang, Jiapeng and Niessner, Matthias and Wonka, Peter},
  journal={ACM Transactions On Graphics (TOG)},
  volume={42},
  number={4},
  pages={1--16},
  year={2023},
}

@inproceedings{long2024wonder3d,
  title={Wonder3d: Single image to 3d using cross-domain diffusion},
  author={Long, Xiaoxiao and Guo, Yuan-Chen and Lin, Cheng and Liu, Yuan and Dou, Zhiyang and Liu, Lingjie and Ma, Yuexin and Zhang, Song-Hai and Habermann, Marc and Theobalt, Christian and others},
  booktitle={CVPR},
  year={2024}
}

@inproceedings{kirillov2023segment,
  title={Segment anything},
  author={Kirillov, Alexander and Mintun, Eric and Ravi, Nikhila and Mao, Hanzi and Rolland, Chloe and Gustafson, Laura and Xiao, Tete and Whitehead, Spencer and Berg, Alexander C and Lo, Wan-Yen and others},
  booktitle={ICCV},
  year={2023}
}

@inproceedings{mo2019partnet,
  title={Partnet: A large-scale benchmark for fine-grained and hierarchical part-level 3d object understanding},
  author={Mo, Kaichun and Zhu, Shilin and Chang, Angel X and Yi, Li and Tripathi, Subarna and Guibas, Leonidas J and Su, Hao},
  booktitle={CVPR},
  year={2019}
}

@inproceedings{lipman2022flow,
  title={Flow matching for generative modeling},
  author={Lipman, Yaron and Chen, Ricky TQ and Ben-Hamu, Heli and Nickel, Maximilian and Le, Matt},
  booktitle={NeurIPS},
  year={2024}
}

@article{yang2024sampart3d,
  title={Sampart3d: Segment any part in 3d objects},
  author={Yang, Yunhan and Huang, Yukun and Guo, Yuan-Chen and Lu, Liangjun and Wu, Xiaoyang and Lam, Edmund Y and Cao, Yan-Pei and Liu, Xihui},
  journal={arXiv preprint arXiv:2411.07184},
  year={2024}
}

@inproceedings{liu2024part123,
  title={Part123: part-aware 3d reconstruction from a single-view image},
  author={Liu, Anran and Lin, Cheng and Liu, Yuan and Long, Xiaoxiao and Dou, Zhiyang and Guo, Hao-Xiang and Luo, Ping and Wang, Wenping},
  booktitle={ACM SIGGRAPH},
  year={2024}
}

@article{chen2024partgen,
  title={PartGen: Part-level 3D Generation and Reconstruction with Multi-View Diffusion Models},
  author={Chen, Minghao and Shapovalov, Roman and Laina, Iro and Monnier, Tom and Wang, Jianyuan and Novotny, David and Vedaldi, Andrea},
  journal={arXiv preprint arXiv:2412.18608},
  year={2024}
}

@article{li2025triposg,
  title={TripoSG: High-Fidelity 3D Shape Synthesis using Large-Scale Rectified Flow Models},
  author={Li, Yangguang and Zou, Zi-Xin and Liu, Zexiang and Wang, Dehu and Liang, Yuan and Yu, Zhipeng and Liu, Xingchao and Guo, Yuan-Chen and Liang, Ding and Ouyang, Wanli and others},
  journal={arXiv preprint arXiv:2502.06608},
  year={2025}
}

@article{xiang2024structured,
  title={Structured 3d latents for scalable and versatile 3d generation},
  author={Xiang, Jianfeng and Lv, Zelong and Xu, Sicheng and Deng, Yu and Wang, Ruicheng and Zhang, Bowen and Chen, Dong and Tong, Xin and Yang, Jiaolong},
  journal={arXiv preprint arXiv:2412.01506},
  year={2024},
}

@article{shi2023zero123plus,
  title={Zero123++: a single image to consistent multi-view diffusion base model},
  author={Shi, Ruoxi and Chen, Hansheng and Zhang, Zhuoyang and Liu, Minghua and Xu, Chao and Wei, Xinyue and Chen, Linghao and Zeng, Chong and Su, Hao},
  journal={arXiv preprint arXiv:2310.15110},
  year={2023}
}

@inproceedings{liu2023syncdreamer,
  title={SyncDreamer: Learning to Generate Multiview-consistent Images from a Single-view Image},
  author={Liu, Yuan and Lin, Cheng and Zeng, Zijiao and Long, Xiaoxiao and Liu, Lingjie and Komura, Taku and Wang, Wenping},
  booktitle={ICLR},
  year={2024}
}

@inproceedings{wang2023score,
  title={Score jacobian chaining: Lifting pretrained 2d diffusion models for 3d generation},
  author={Wang, Haochen and Du, Xiaodan and Li, Jiahao and Yeh, Raymond A and Shakhnarovich, Greg},
  booktitle={CVPR},
  year={2023}
}

@inproceedings{wang2023prolificdreamer,
  title={Prolificdreamer: High-fidelity and diverse text-to-3d generation with variational score distillation},
  author={Wang, Zhengyi and Lu, Cheng and Wang, Yikai and Bao, Fan and Li, Chongxuan and Su, Hang and Zhu, Jun},
  booktitle={NeurIPS},
  year={2023}
}

@inproceedings{lin2023magic3d,
  title={Magic3d: High-resolution text-to-3d content creation},
  author={Lin, Chen-Hsuan and Gao, Jun and Tang, Luming and Takikawa, Towaki and Zeng, Xiaohui and Huang, Xun and Kreis, Karsten and Fidler, Sanja and Liu, Ming-Yu and Lin, Tsung-Yi},
  booktitle={CVPR},
  year={2023}
}

@inproceedings{huang2023dreamtime,
  title={DreamTime: An Improved Optimization Strategy for Text-to-3D Content Creation},
  author={Huang, Yukun and Wang, Jianan and Shi, Yukai and Qi, Xianbiao and Zha, Zheng-Jun and Zhang, Lei},
  booktitle={ICLR},
  year={2024}
}

@inproceedings{chen2023fantasia3d,
  title={Fantasia3d: Disentangling geometry and appearance for high-quality text-to-3d content creation},
  author={Chen, Rui and Chen, Yongwei and Jiao, Ningxin and Jia, Kui},
  booktitle={ICCV},
  year={2023}
}

@inproceedings{deitke2022objaverse,
  title={Objaverse: A universe of annotated 3d objects},
  author={Deitke, Matt and Schwenk, Dustin and Salvador, Jordi and Weihs, Luca and Michel, Oscar and VanderBilt, Eli and Schmidt, Ludwig and Ehsani, Kiana and Kembhavi, Aniruddha and Farhadi, Ali},
  booktitle={CVPR},
  year={2023}
}

@inproceedings{metzer2022latentnerf,
  title={Latent-nerf for shape-guided generation of 3d shapes and textures},
  author={Metzer, Gal and Richardson, Elad and Patashnik, Or and Giryes, Raja and Cohen-Or, Daniel},
  booktitle={CVPR},
  year={2023}
}

@inproceedings{yang2024dreamcomposer,
  title={{DreamComposer: Controllable 3D Object Generation via Multi-View Conditions}},
  author={Yang, Yunhan and Huang, Yukun and Wu, Xiaoyang and Guo, Yuan-Chen and Zhang, Song-Hai and Zhao, Hengshuang and He, Tong and Liu, Xihui},
  booktitle={CVPR},
  year={2024}
}

@article{xu2024instantmesh,
  title={Instantmesh: Efficient 3d mesh generation from a single image with sparse-view large reconstruction models},
  author={Xu, Jiale and Cheng, Weihao and Gao, Yiming and Wang, Xintao and Gao, Shenghua and Shan, Ying},
  journal={arXiv preprint arXiv:2404.07191},
  year={2024}
}

@inproceedings{liu2023one2345,
  title={One-2-3-45++: Fast Single Image to 3D Objects with Consistent Multi-View Generation and 3D Diffusion}, 
  author={Minghua Liu and Ruoxi Shi and Linghao Chen and Zhuoyang Zhang and Chao Xu and Xinyue Wei and Hansheng Chen and Chong Zeng and Jiayuan Gu and Hao Su},
  booktitle={CVPR},
  year={2024},
}

@inproceedings{yi2023gaussiandreamer,
  title={Gaussiandreamer: Fast generation from text to 3d gaussian splatting with point cloud priors},
  author={Yi, Taoran and Fang, Jiemin and Wu, Guanjun and Xie, Lingxi and Zhang, Xiaopeng and Liu, Wenyu and Tian, Qi and Wang, Xinggang},
  booktitle={CVPR},
  year={2024}
}

@inproceedings{wu2024consistent3d,
  title={Consistent3d: Towards consistent high-fidelity text-to-3d generation with deterministic sampling prior},
  author={Wu, Zike and Zhou, Pan and Yi, Xuanyu and Yuan, Xiaoding and Zhang, Hanwang},
  booktitle={CVPR},
  year={2024}
}

@inproceedings{alldieck2024score,
  title={Score Distillation Sampling with Learned Manifold Corrective}, 
  author={Thiemo Alldieck and Nikos Kolotouros and Cristian Sminchisescu},
  booktitle={ECCV},
  year={2024},
}

@article{tang2023stable,
  title={Stable Score Distillation for High-Quality 3D Generation}, 
  author={Boshi Tang and Jianan Wang and Zhiyong Wu and Lei Zhang},
  journal={arXiv preprint: 2312.09305},
  year={2023},
}

@inproceedings{yan2024DreamView,
  title={DreamView: Injecting View-specific Text Guidance into Text-to-3D Generation},
  author={Yan, Junkai and Gao, Yipeng and Yang, Qize and Wei, Xihan and Xie, Xuansong and Wu, Ancong and Zheng, Wei-Shi},
  booktitle={ECCV},
  year={2024},
}

@article{qi2024tailor3d,
  title={Tailor3d: Customized 3d assets editing and generation with dual-side images},
  author={Qi, Zhangyang and Yang, Yunhan and Zhang, Mengchen and Xing, Long and Wu, Xiaoyang and Wu, Tong and Lin, Dahua and Liu, Xihui and Wang, Jiaqi and Zhao, Hengshuang},
  journal={arXiv preprint arXiv:2407.06191},
  year={2024}
}

@inproceedings{zou2024triplane,
  title={Triplane meets gaussian splatting: Fast and generalizable single-view 3d reconstruction with transformers},
  author={Zou, Zi-Xin and Yu, Zhipeng and Guo, Yuan-Chen and Li, Yangguang and Liang, Ding and Cao, Yan-Pei and Zhang, Song-Hai},
  booktitle={CVPR},
  year={2024}
}

@inproceedings{chen2024comboverse,
  title={Comboverse: Compositional 3d assets creation using spatially-aware diffusion guidance},
  author={Chen, Yongwei and Wang, Tengfei and Wu, Tong and Pan, Xingang and Jia, Kui and Liu, Ziwei},
  booktitle={ECCV},
  year={2024},
}

@article{liu2025partfield,
  title={PARTFIELD: Learning 3D Feature Fields for Part Segmentation and Beyond},
  author={Liu, Minghua and Uy, Mikaela Angelina and Xiang, Donglai and Su, Hao and Fidler, Sanja and Sharp, Nicholas and Gao, Jun},
  journal={arXiv preprint arXiv:2504.11451},
  year={2025}
}

@article{li2024pasta,
  title={PASTA: Controllable Part-Aware Shape Generation with Autoregressive Transformers},
  author={Li, Songlin and Paschalidou, Despoina and Guibas, Leonidas},
  journal={arXiv preprint arXiv:2407.13677},
  year={2024}
}

@article{yan2024phycage,
  title={PhyCAGE: Physically Plausible Compositional 3D Asset Generation from a Single Image},
  author={Yan, Han and Zhang, Mingrui and Li, Yang and Ma, Chao and Ji, Pan},
  journal={arXiv preprint arXiv:2411.18548},
  year={2024}
}

@article{tang2025efficient,
  title={Efficient Part-level 3D Object Generation via Dual Volume Packing},
  author={Tang, Jiaxiang and Lu, Ruijie and Li, Zhaoshuo and Hao, Zekun and Li, Xuan and Wei, Fangyin and Song, Shuran and Zeng, Gang and Liu, Ming-Yu and Lin, Tsung-Yi},
  journal={arXiv preprint arXiv:2506.09980},
  year={2025}
}

@article{lin2025partcrafter,
  title={PartCrafter: Structured 3D Mesh Generation via Compositional Latent Diffusion Transformers},
  author={Lin, Yuchen and Lin, Chenguo and Pan, Panwang and Yan, Honglei and Feng, Yiqiang and Mu, Yadong and Fragkiadaki, Katerina},
  journal={arXiv preprint arXiv:2506.05573},
  year={2025}
}

@inproceedings{huang2024epidiff,
  title={Epidiff: Enhancing multi-view synthesis via localized epipolar-constrained diffusion},
  author={Huang, Zehuan and Wen, Hao and Dong, Junting and Wang, Yaohui and Li, Yangguang and Chen, Xinyuan and Cao, Yan-Pei and Liang, Ding and Qiao, Yu and Dai, Bo and others},
  booktitle={CVPR},
  year={2024}
}

@inproceedings{wen2025ouroboros3d,
  title={Ouroboros3d: Image-to-3d generation via 3d-aware recursive diffusion},
  author={Wen, Hao and Huang, Zehuan and Wang, Yaohui and Chen, Xinyuan and Sheng, Lu},
  booktitle={CVPR},
  year={2025}
}

@article{yang2025omnipart,
  title={Omnipart: Part-aware 3d generation with semantic decoupling and structural cohesion},
  author={Yang, Yunhan and Zhou, Yufan and Guo, Yuan-Chen and Zou, Zi-Xin and Huang, Yukun and Liu, Ying-Tian and Xu, Hao and Liang, Ding and Cao, Yan-Pei and Liu, Xihui},
  journal={arXiv preprint arXiv:2507.06165},
  year={2025}
}

@inproceedings{sun2024generative,
  title={Generative multimodal models are in-context learners},
  author={Sun, Quan and Cui, Yufeng and Zhang, Xiaosong and Zhang, Fan and Yu, Qiying and Wang, Yueze and Rao, Yongming and Liu, Jingjing and Huang, Tiejun and Wang, Xinlong},
  booktitle={CVPR},
  year={2024}
}

@article{chen2025blip3,
  title={Blip3-o: A family of fully open unified multimodal models-architecture, training and dataset},
  author={Chen, Jiuhai and Xu, Zhiyang and Pan, Xichen and Hu, Yushi and Qin, Can and Goldstein, Tom and Huang, Lifu and Zhou, Tianyi and Xie, Saining and Savarese, Silvio and others},
  journal={arXiv preprint arXiv:2505.09568},
  year={2025}
}

@article{bai2025qwen2,
  title={Qwen2. 5-vl technical report},
  author={Bai, Shuai and Chen, Keqin and Liu, Xuejing and Wang, Jialin and Ge, Wenbin and Song, Sibo and Dang, Kai and Wang, Peng and Wang, Shijie and Tang, Jun and others},
  journal={arXiv preprint arXiv:2502.13923},
  year={2025}
}

@article{cao2025physx3d,
  title={Physx-3d: Physical-grounded 3d asset generation},
  author={Cao, Ziang and Chen, Zhaoxi and Pan, Liang and Liu, Ziwei},
  journal={arXiv preprint arXiv:2507.12465},
  year={2025}
}

@article{wang2025embodiedgen,
  title={Embodiedgen: Towards a generative 3d world engine for embodied intelligence},
  author={Wang, Xinjie and Liu, Liu and Cao, Yu and Wu, Ruiqi and Qin, Wenkang and Wang, Dehui and Sui, Wei and Su, Zhizhong},
  journal={arXiv preprint arXiv:2506.10600},
  year={2025}
}

@article{chen2025autopartgen,
  title={Autopartgen: Autogressive 3d part generation and discovery},
  author={Chen, Minghao and Wang, Jianyuan and Shapovalov, Roman and Monnier, Tom and Jung, Hyunyoung and Wang, Dilin and Ranjan, Rakesh and Laina, Iro and Vedaldi, Andrea},
  journal={arXiv preprint arXiv:2507.13346},
  year={2025}
}

@inproceedings{dong2025one,
  title={From one to more: Contextual part latents for 3d generation},
  author={Dong, Shaocong and Ding, Lihe and Chen, Xiao and Li, Yaokun and Wang, Yuxin and Wang, Yucheng and Wang, Qi and Kim, Jaehyeok and Gao, Chenjian and Huang, Zhanpeng and others},
  booktitle={ICCV},
  year={2025}
}

@inproceedings{ye2024dreamreward,
  title={Dreamreward: Text-to-3d generation with human preference},
  author={Ye, Junliang and Liu, Fangfu and Li, Qixiu and Wang, Zhengyi and Wang, Yikai and Wang, Xinzhou and Duan, Yueqi and Zhu, Jun},
  booktitle={ECCV},
  year={2024},
}

@article{liu2025dreamreward,
  title={Dreamreward-x: Boosting high-quality 3d generation with human preference alignment},
  author={Liu, Fangfu and Ye, Junliang and Wang, Yikai and Wang, Hanyang and Wang, Zhengyi and Zhu, Jun and Duan, Yueqi},
  journal={TPAMI},
  year={2025},
}

@inproceedings{wang2024animatabledreamer,
  title={Animatabledreamer: Text-guided non-rigid 3d model generation and reconstruction with canonical score distillation},
  author={Wang, Xinzhou and Wang, Yikai and Ye, Junliang and Sun, Fuchun and Wang, Zhengyi and Wang, Ling and Liu, Pengkun and Sun, Kai and Wang, Xintong and Xie, Wende and others},
  booktitle={ECCV},
  year={2024},
}

@inproceedings{liu2023paris,
  title={Paris: Part-level reconstruction and motion analysis for articulated objects},
  author={Liu, Jiayi and Mahdavi-Amiri, Ali and Savva, Manolis},
  booktitle={ICCV},
  year={2023}
}

@inproceedings{liu2025building,
  title={Building interactable replicas of complex articulated objects via gaussian splatting},
  author={Liu, Yu and Jia, Baoxiong and Lu, Ruijie and Ni, Junfeng and Zhu, Song-Chun and Huang, Siyuan},
  booktitle={ICLR},
  year={2025}
}

@inproceedings{weng2024neural,
  title={Neural implicit representation for building digital twins of unknown articulated objects},
  author={Weng, Yijia and Wen, Bowen and Tremblay, Jonathan and Blukis, Valts and Fox, Dieter and Guibas, Leonidas and Birchfield, Stan},
  booktitle={CVPR},
  year={2024}
}

@article{wu2025reartgs,
  title={Reartgs: Reconstructing and generating articulated objects via 3d gaussian splatting with geometric and motion constraints},
  author={Wu, Di and Liu, Liu and Linli, Zhou and Huang, Anran and Song, Liangtu and Yu, Qiaojun and Wu, Qi and Lu, Cewu},
  journal={arXiv preprint arXiv:2503.06677},
  year={2025}
}

@inproceedings{song2024reacto,
  title={Reacto: Reconstructing articulated objects from a single video},
  author={Song, Chaoyue and Wei, Jiacheng and Foo, Chuan Sheng and Lin, Guosheng and Liu, Fayao},
  booktitle={CVPR},
  year={2024}
}

@inproceedings{tu2025dreamo,
  title={Dreamo: Articulated 3d reconstruction from a single casual video},
  author={Tu, Tao and Li, Ming-Feng and Lin, Chieh Hubert and Cheng, Yen-Chi and Sun, Min and Yang, Ming-Hsuan},
  booktitle={WACV},
  year={2025},
}

@article{chen2024urdformer,
  title={Urdformer: A pipeline for constructing articulated simulation environments from real-world images},
  author={Chen, Zoey and Walsman, Aaron and Memmel, Marius and Mo, Kaichun and Fang, Alex and Vemuri, Karthikeya and Wu, Alan and Fox, Dieter and Gupta, Abhishek},
  journal={arXiv preprint arXiv:2405.11656},
  year={2024}
}

@inproceedings{gao2025meshart,
  title={MeshArt: Generating Articulated Meshes with Structure-guided Transformers},
  author={Gao, Daoyi and Siddiqui, Yawar and Li, Lei and Dai, Angela},
  booktitle={CVPR},
  year={2025}
}

@article{le2024articulate,
  title={Articulate-anything: Automatic modeling of articulated objects via a vision-language foundation model},
  author={Le, Long and Xie, Jason and Liang, William and Wang, Hung-Ju and Yang, Yue and Ma, Yecheng Jason and Vedder, Kyle and Krishna, Arjun and Jayaraman, Dinesh and Eaton, Eric},
  journal={arXiv preprint arXiv:2410.13882},
  year={2024}
}

@inproceedings{liu2024cage,
  title={Cage: Controllable articulation generation},
  author={Liu, Jiayi and Tam, Hou In Ivan and Mahdavi-Amiri, Ali and Savva, Manolis},
  booktitle={CVPR},
  year={2024}
}

@article{liu2024singapo,
  title={Singapo: Single image controlled generation of articulated parts in objects},
  author={Liu, Jiayi and Iliash, Denys and Chang, Angel X and Savva, Manolis and Mahdavi-Amiri, Ali},
  journal={arXiv preprint arXiv:2410.16499},
  year={2024}
}

@article{mandi2024real2code,
  title={Real2code: Reconstruct articulated objects via code generation},
  author={Mandi, Zhao and Weng, Yijia and Bauer, Dominik and Song, Shuran},
  journal={arXiv preprint arXiv:2406.08474},
  year={2024}
}

@article{qiu2025articulate,
  title={Articulate anymesh: Open-vocabulary 3d articulated objects modeling},
  author={Qiu, Xiaowen and Yang, Jincheng and Wang, Yian and Chen, Zhehuan and Wang, Yufei and Wang, Tsun-Hsuan and Xian, Zhou and Gan, Chuang},
  journal={arXiv preprint arXiv:2502.02590},
  year={2025}
}

@inproceedings{zhan2024oakink2,
  title={Oakink2: A dataset of bimanual hands-object manipulation in complex task completion},
  author={Zhan, Xinyu and Yang, Lixin and Zhao, Yifei and Mao, Kangrui and Xu, Hanlin and Lin, Zenan and Li, Kailin and Lu, Cewu},
  booktitle={CVPR},
  year={2024}
}

@inproceedings{xiang2020sapien,
  title={Sapien: A simulated part-based interactive environment},
  author={Xiang, Fanbo and Qin, Yuzhe and Mo, Kaichun and Xia, Yikuan and Zhu, Hao and Liu, Fangchen and Liu, Minghua and Jiang, Hanxiao and Yuan, Yifu and Wang, He and others},
  booktitle={CVPR},
  year={2020}
}

@article{lian2025infinite,
  title={Infinite Mobility: Scalable High-Fidelity Synthesis of Articulated Objects via Procedural Generation},
  author={Lian, Xinyu and Yu, Zichao and Liang, Ruiming and Wang, Yitong and Luo, Li Ray and Chen, Kaixu and Zhou, Yuanzhen and Tang, Qihong and Xu, Xudong and Lyu, Zhaoyang and others},
  journal={arXiv preprint arXiv:2503.13424},
  year={2025}
}

@InProceedings{Mu_2025_CVPR,
    author    = {Mu, Yao and Chen, Tianxing and Chen, Zanxin and Peng, Shijia and Lan, Zhiqian and Gao, Zeyu and Liang, Zhixuan and Yu, Qiaojun and Zou, Yude and Xu, Mingkun and Lin, Lunkai and Xie, Zhiqiang and Ding, Mingyu and Luo, Ping},
    title     = {RoboTwin: Dual-Arm Robot Benchmark with Generative Digital Twins},
    booktitle = {CVPR},
    year      = {2025},
}

@article{chen2025robotwin,
  title={Robotwin 2.0: A scalable data generator and benchmark with strong domain randomization for robust bimanual robotic manipulation},
  author={Chen, Tianxing and Chen, Zanxin and Chen, Baijun and Cai, Zijian and Liu, Yibin and Li, Zixuan and Liang, Qiwei and Lin, Xianliang and Ge, Yiheng and Gu, Zhenyu and others},
  journal={arXiv preprint arXiv:2506.18088},
  year={2025}
}

@article{ding2025fullpart,
  title={FullPart: Generating each 3D Part at Full Resolution},
  author={Ding, Lihe and Dong, Shaocong and Li, Yaokun and Gao, Chenjian and Chen, Xiao and Han, Rui and Kuang, Yihao and Zhang, Hong and Huang, Bo and Huang, Zhanpeng and others},
  journal={arXiv preprint arXiv:2510.26140},
  year={2025}
}

@article{cao2025physx,
  title={PhysX-Anything: Simulation-Ready Physical 3D Assets from Single Image},
  author={Cao, Ziang and Hong, Fangzhou and Chen, Zhaoxi and Pan, Liang and Liu, Ziwei},
  journal={arXiv preprint arXiv:2511.13648},
  year={2025}
}

@article{he2025unipart,
  title={UniPart: Part-Level 3D Generation with Unified 3D Geom-Seg Latents},
  author={He, Xufan and Wu, Yushuang and Guo, Xiaoyang and Ye, Chongjie and Zhou, Jiaqing and Hu, Tianlei and Han, Xiaoguang and Du, Dong},
  journal={arXiv preprint arXiv:2512.09435},
  year={2025}
}

@article{lai2025lattice,
  title={LATTICE: Democratize High-Fidelity 3D Generation at Scale},
  author={Lai, Zeqiang and Zhao, Yunfei and Zhao, Zibo and Liu, Haolin and Lin, Qingxiang and Huang, Jingwei and Guo, Chunchao and Yue, Xiangyu},
  journal={arXiv preprint arXiv:2512.03052},
  year={2025}
}

@article{zhao2023michelangelo,
  title={Michelangelo: Conditional 3d shape generation based on shape-image-text aligned latent representation},
  author={Zhao, Zibo and Liu, Wen and Chen, Xin and Zeng, Xianfang and Wang, Rui and Cheng, Pei and Fu, Bin and Chen, Tao and Yu, Gang and Gao, Shenghua},
  journal={NeurIPS},
  year={2023}
}

\newpage
\appendix
\onecolumn

\end{document}